\documentclass[a4paper,twoside]{article}
\usepackage{epsfig}
\usepackage{subcaption}
\usepackage{calc}
\usepackage{amssymb}
\usepackage{amstext}
\usepackage{amsmath}
\usepackage{amsthm}
\usepackage{multicol}
\usepackage{pslatex}
\usepackage{apalike}
\usepackage{hyperref}
\usepackage{enumitem}
\usepackage[bottom]{footmisc}
\usepackage{xcolor}
\usepackage{lmodern}

\usepackage[T1]{fontenc}

\usepackage{textcomp}
\usepackage{SCITEPRESS}
\usepackage{etoolbox}

\begin{document}

\title{The Human-or-Machine Matter:\\ Turing-Inspired Reflections on an Everyday Issue}

\author{\authorname{David Harel and Assaf Marron}
\vspace{0.2cm}
 \affiliation{Department of Computer Science and Applied Mathematics, Weizmann Institute of Science, \\ Rehovot, 76100, Israel}
\email{dharel@weizmann.ac.il, assaf.marron@weizmann.ac.il} 
}


\abstract{
In his seminal paper ``Computing Machinery and Intelligence'', Alan Turing introduced the ``imitation game'' as part of exploring the concept of machine intelligence. The Turing Test 
has since been the subject of much analysis, debate, refinement and extension. 
Here we sidestep the question of whether a particular machine can be labeled intelligent, or can be said to match  human capabilities in a given context. 
Instead,
we first draw attention to the seemingly simpler question a person may ask themselves in an everyday interaction: ``Am I interacting with a human or with a machine?''. We then shift the focus from seeking a method for eliciting the answer, and, rather,  reflect  
upon the importance 
and significance 
of this Human-or-Machine question and the use one may make of a reliable answer thereto. 
Whereas Turing's original test is widely considered to be more of a thought experiment, 
the Human-or-Machine matter
as discussed here has obvious practical 
relevance.
While it is still unclear if and when machines will be able to mimic human behavior with high fidelity in everyday contexts, we argue that near-term exploration of the issues raised here can contribute to 
refinement of methods for developing computerized systems, 
and may also lead to new insights into fundamental characteristics of human behavior.
}

\maketitle

\section {Introduction}\label{sec:introduction}
Turing's 1950 paper~\cite{turing1950} introduced the famed ``imitation game'' as part of the definition of the concept of machine intelligence; i.e., as a means to determine whether a computer can be labeled intelligent. 
Over the years, the Turing test has been the subject of much work, and has resulted in several variants ~\cite{french2000turing50Years,hoffmann2022ai70years}.  
Similar tests have also been proposed in quite different contexts ~\cite{hagen2023turingTestForModel,harel2005turingTestBio,beven2022validatingHydrologyModelsTuring,harel2016niepceBellOdorTuring,kalik2007automotiveTuringTest,rosemarinRosenfeld2019playingChess,bush2020turingMoleculeGenerator}. 

Here, we completely sidestep the conceptual, philosophical
issue of defining or measuring intelligence, as well as the practical  question of whether a machine can be built to replace, or mimic, a person in the performance of some
specific task~\cite{sifakis2023testingMachineIntelligence}.

Instead, we look more broadly at a concept that we term \textit{The Human-or-Machine Matter}. In a future world, when in some interactions machines will be able to mimic humans impressively, new social, psychological, functional and technical issues are bound to become relevant. For example: 
 \begin{enumerate}
    \item Will humans care whether the agent they interact with is a human or a machine, and if yes, why? 
    \item How will a person's behavior or emotional state differ between interactions with another human and interactions with a machine whose behavior is indistinguishable from a human's?
    \item How will the answer to the question of an agent's human-or-machine identity (hereafter, the {\emph{H-or-M question})} be elicited? 
       \item Will human language and social practices change in such cases? 
       \item Will machine-machine interactions change when the behavior of one or both of the participants is very close to a human's?
     \item will machines indeed be indistinguishable  from  humans, or will this be a non-issue because openly taking advantage of machine capabilities will be prioritized over manifesting human-like behavior? 
\end{enumerate}

In examining these issues, we discuss research, opinions and predictions about differences between humans and machines, and differences between human-human and human-machine interactions.

Of special interest here is the importance of the \textit{Human-or-Machine question}, which stands for the interest, or curiosity, of a person who is engaged in an everyday interaction  with an agent, and is wondering whether the agent is a human or a machine.  

While in general, machines are at present unable to perform non-trivial interactive tasks in a way that conceals the fact that they are machines and not humans, we expect that in the future this will change dramatically.  
This will be supported by a combination of the growing presence of  of machines performing functions that were traditionally carried out by humans, as in 
service centers with automated chatbots, 
healthcare conversational agents, service robots in stores, 
autonomous vehicles~\cite{sheth2019cognitiveChatbotsInService,parmar2022healthConversationalAgentsNature,frank2023relationsWithRobotService,bbc2021AV,sadeghian2022artificialColleague}, and the prevalence of interactions that hide the agent identity as in text-only or voice-only interactions, or when not being able to see the driver of another vehicle.   

Even when the agent with whom we are interacting is known to be a human, an appropriate variant of the question is still relevant, since that human's behavior may be dictated by a machine, as in a person reading aloud some text composed by a machine or driving a car by merely following computer-generated 
operation and
navigation instructions.

Similarly, when the agent is conspicuously a machine, the H-or-M question asks who is behind the \emph{decisions} or \emph{choices} underlying the agent's role in the interaction. For example, when the agent displays a piece of text as part of an interaction, we want to know if the act of composing  the text and the decision to display it right now were carried out by a human or by the agent.  

If one insists, the original Turing test and its many variants can be viewed as a special case of dealing with the H-or-M question, where, in a controlled laboratory setting,
a human acts as an interrogator in attempting to reveal the human-or-machine identity of an unidentified and hidden agent, based on observing the agent's responses and actions to his or her prodding questions. 
Our inspiration from the Turing test here is manifested in our focus on confined human-agent interactions, rather than on the broader issues of the role of new human-like machines in the world, or on forensic issues like whether a stand-alone, non-interactive artifact, like a piece of text, a picture, a video clip, or a medical diagnosis, was created by a person or a machine~\cite{noll1966humanOrMachineMondrian}. 
Still, here we diverge from classical treatments of the Turing test in that: 
(i) we concentrate on everyday interactions, rather than on a controlled lab setup; (ii) we are interested in how the agent's H-or-M identity affects the interaction at hand; 
(iii) the question of whether it is possible at all to extract the agent's H-or-M identity from a sequence of exchanges is secondary, and we only briefly discuss here issues related to interrogation techniques for extracting this answer. 
(iv) we are not interested in  the question of whether the agent can be labeled as being intelligent.

Parts of our discussion are presented as  questions, some of which
may justify separate, focused research efforts by scientists, philosophers, etc.

The Human-or-Machine issue is presented here as a binary one. Clearly, there may be mixed modes. For example: (i)  the apparent agent is a human physician, who, while consulting a human patient, relies extensively on online search for information or is informed, openly or discreetly, by an automated agent listening in on the conversation; (ii) the apparent agent is a vehicle, but while in many aspects it acts autonomously, it is also remotely supervised and occasionally even controlled by humans (and we are interested in this vehicle's interactions with other road users); (iii) setups like the  above two, but the mixed-mode agent depends on more than one machine and/or more than one human; (iv) extending the above one-to-one human-agent interactions to group interactions, and potentially even without a clear delineation of ``the agent''. Such mixed-mode agents may be treated by default in the same way as a pure machine or a pure human. We defer to future work discussion of the cases where the mixed mode is substantially different from the binary one. 

\section {Are We Different When We Interact with Machines?}\label{sec:HorMrelevance}
One kind of relevance the H-or-M question might have, is to the way in which knowing the answer could affect human behavior. 

The relationship between actual humans and machines that present themselves as almost human, has been explored in a variety of ways in arts, science and philosophy. Suffices to consider movies like ``The Matrix'', ``Blade Runner'', ``The Terminator'' series, and ``Her'',  and books like  ``Machines Like Me'',  and ``I, Robot''\footnote{Excluded from this paper is a comprehensive summary of how these movies and books present the relations and interactions between humans and human-like machine agents, which we were able to readily obtain with a few queries to OpenAI's ChatGPT.}. Scientists have also researched human-machine relations~\cite{chaturvedi2023socialCompanionAI,reinkemeier2022match,milcent2022facial,hindriks2022effectsofClothing,heiser1979paranoyaSimulation}
and proposed that the science of sociology should study AI-related issues~\cite{woolgar1985SociologyAndAI,liu2021sociologicalPerspectives}.

There are many studies of human interactions with chatbots---text-based conversational agents; see, for example,  literature reviews in~\cite{rapp2021humanChatbotInteraction,chaturvedi2023socialCompanionAI,nicolescu2022HCIcustomerService,pinxteren2020ConversationalAgents,mariani2023AIConversationalAgents} and references therein. Research themes include the analysis of chatbot functionality and its relationship to certain success factors, such as the ability to affect user actions; aspects of the interactions, such as the language used or length of conversation; and human-chatbot relations, such as acceptance and trust. 
Studies that  focus specifically on the differences between humans and human-like machines  in normal kinds of interaction contexts (like a service robot in a store) are  also emerging\cite{frank2023relationsWithRobotService}. In most of that work 
the H-or-M question itself is not at the center of the research. In many, the fact that the agent is a machine is disclosed up front, and in others the researchers were interested in whether or not the human users ascribed humanness, or human-like behavior, to a machine agent. 

Below are some examples of person-agent interactions, where having an answer may affect the person's behavior, even in everyday routine interactions. 
In this discussion, one should note, though, that while validating or refuting each such candidate effect on the user's actions or emotional state is an intriguing issue, the examples appear here only to support the main claim of the paper; i.e., that the H-or-M matter will fast become relevant in many everyday contexts. 

Some languages require distinguishing humans from  non-human agents, and in the case of a human agent often also detecting the gender. A person conducting a text exchange with a service center may be inclined to use different pronouns or verbs for humans and for machines, both when addressing the representative and when discussing what another representative may have communicated in a prior exchange. Furthermore, special linguistic patterns may evolve for cases where such a determination remains unknown.

In order to plan his or her next steps, a person is more likely to try to figure out particular patterns in the behavior of the conversing agent if the agent were known to be a machine, rather than a human, and to make more of an effort to relate to those patterns. One reason for this is that machines are expected to follow patterns more consistently than humans. Also, when we know that we are interacting with a machine, we may be able to find out more information about the machine, like its make and model, the software controlling it, etc., and then use shared knowledge about what to expect or how to interact with 
this particular type of agent.  

As summarized in~\cite{rapp2021humanChatbotInteraction},
some research on human-chatbot interactions as compared with human-human ones suggests that when interacting with a machine the human may be briefer and less polite, more inclined to abruptly stop or divert a conversation, or even employ profanities~\cite{hill2015realConversations}. 
It is not clear if this is because we do not expect the machine to be really offended, or because we are not concerned that our impolite behavior might bring about  repercussions.

One may wonder if we will be more accepting of a machine agent's
formal, dry, or even rude attitude, knowing that machines will not normally be considerate.  Similarly, will people be more patient with "stupid" or repeated answers, or 
with inconsiderate actions, such as when driving behind an overly cautious and slow autonomous vehicle (AV), knowing that machines are limited, and their behavior cannot be readily changed?
We expect that people will be less patient when experiencing delayed responses, knowing that machines are usually fast and task oriented.

The present emphasis on prompt writing and prompt engineering skills for interacting with Large Language Models (LLMs) suggests that we will make a stronger effort to  explain ourselves, knowing that a machine is expected to be more limited than a human in understanding our intentions and needs. 
Also, one can expect that we 
will be more inclined to report a machine's dangerous or undesired behavior to the machine itself, or to its owner or manufacturer, or even to the police, seeing our actions as desired feedback and expecting more productive reactions than when criticizing or reporting a person.  

The issues of trust building, the willingness to disclose personal information, developing a personal relationship with and feeling empathy towards machine agents, have all been discussed in the literature~\cite{rapp2021humanChatbotInteraction,chaturvedi2023socialCompanionAI,pinxteren2020ConversationalAgents,mariani2023AIConversationalAgents,cai2022anthropomorphism}. 
Some research shows difficulties in these areas, 
which may be partly related to the agents being perceived as uncanny. Other research has shown a much ``warmer'' attitude by users. It is yet to be determined how this aspect will evolve 
as the technology advances and becomes more pervasive. 

Will we be more open to learning from the machine's behavior? 
Consider for example the following: 
When observing an AV negotiating a certain class of complicated scenarios differently from the way we would have dealt with it, 
we might be inclined to mimic the AV, assuming that much thought and serious design and testing had been carried out to yield such behavior.  However, even that is far from obvious, and we might actually do quite the opposite: unlike the natural tendency to ``follow the crowd'', we may prefer to make our own decisions in such cases, thinking  ourselves to be ``smarter'' and more experienced than a typical machine.
    
It would also be interesting to identify areas in which having the answer to the H-or-M question  does \emph{not} affect human behavior. Would we still be curious about the answer? Will the question arise subconsciously, like the inevitable tendency to try to incorporate gender perception in first impressions
\cite{signorella1992rememberingGender,ruble1986stalkingGender}? 
What purpose would this curiosity or knowledge serve? And if such curiosity does not emerge,  will the indifference to  whether the agent with whom we are interacting is a human or a machine affect overall patterns of social interaction in the relevant area?

Another facet of the H-or-M question is the effect of an incorrect determination. For example, will a human agent be offended when they realize that a person they are interacting with thinks they are a machine? How will that person feel when they realize their mistake? 
How embarrassed or angry will a person be when they realize that the agent (perhaps even a co-worker~\cite{sadeghian2022artificialColleague}) whom they thought was human, and with whom they have developed a relationship, is a machine?    

\section {When and How Should H-or-M be Easily Resolvable?}\label{sec:HorMIndications}

Given the relevance of the H-or-M identity of an agent, when and how should this information be made readily available. 

In what contexts should the answer be provided to the person once, explicitly,  
either in advance, as is the case with some service chatbots, or perhaps
constantly and automatically, as is done  with  "recording in progress" indicators in phone calls and teleconferences? 

Presently most chatbots disclose the fact that they are machines. Should autonomous vehicles be clearly marked as such? Should autonomous drones be marked differently from remotely controlled ones~\cite{traboulsi2021HorMforDrone}?
Should a human-like receptionist robot be clearly marked as such, in order to not be mistakenly thought to be human?  
And should interactions with  human agents be labeled as such, or should this be the default?

Should there be standards for communicating this information --- using, say, text, icons or spoken words? In what cases should the information be provided in response to programming interfaces? 

When should the H-or-M question be left for the interested person to answer for themselves, without a dedicated, explicit interface? 
One context in which this is likely to be the case is when the agent's behavior is clearly a mixed-mode as described above , a collaborative operation,  partly human and partly machine. The exact  division of subtasks may be interesting to humans, but may not be readily available.  

Are there cases where the answer should actually be hidden? We are interested here mainly in ethical contexts, excluding situations of 
conflict, oppression or fraud.
Thus, for example, detecting whether interactions in social media or in dating websites are with a human or with a machine is outside the scope of this paper. 
Note that in these excluded cases, the issues may be not just with regard to the H-or-M identity of the agent, but more general, as 
 when agents are not really who they claim to be. 
Hence, we ask, are there ethical cases where even searching for the answer should be forbidden or blocked? Here are a few candidate scenarios: 
\begin{itemize}
    \item When the agent's role is to help train a human user in interacting with other humans, complete with their errors and misunderstandings,  as in the training of a therapist, a dancer or a sports person. 
    \item When the agent is part of a show, or a meta-discussion, where the enigma of the H-or-M identity of the agent is part of the essence of the interaction. 
    \item When during development of a system that needs to mimic human interactions with humans, the system is trained or tested through interaction with a person who works under the assumption that they are interacting with another human. 
    \item A variety of research situations studying the behaviors of humans and machines. 
\end{itemize}

\section {Some Inherent Differences Between Humans and Machines} \label{sec:distinctions}

One cannot delve into the H-or-M matter without considering the essential differences between the behavior of human agents and that of machine agents---in general and in specific contexts. 
Turing himself dedicated a section to such a discussion in his 1950 paper~\cite{turing1950}, though clearly some distinctions  have changed dramatically over time; e.g., in the capability to learn and to adapt to changing conditions.
Such differences between humans and machines are sometimes phrased as goals in achieving artificial intelligence in perception, cognition and reasoning (see, for example,~\cite[Ch.~6,7]{sifakis2022understandingTheWorld} and~\cite[Ch.~1.1]{russellNorvig2002AI} and references therein) and in achieving a sense of humanness when interacting with machine agents~\cite{rapp2021humanChatbotInteraction,setlur2022chatbotDesign,cai2022anthropomorphism}

Gaining insights into these inherent differences can help in studying their effects on interactions and in designing interrogation strategies.  

Without attempting to get into 
a deep psychological, biological and philosophical discussion, we list below some such tentative differences, as they may be identified by typical 
people in the context of everyday interactions. 
Our intention is to draw attention to the existence of  
behavioral differences between humans and machines 
in interaction, and to invite discussion of their consequences. 

\begin{itemize}

\item  Free will: ``Machines are completely pre-programmed, whereas humans have free will.''  

\item  Emotions: ``Humans have emotions and feel compassion, whereas  machines do not.''

\item  Context awareness: ``Humans are sensitive to context and to innumerable explicit and tacit inputs, to which a typical machine is blind.'' 

\item  Common sense and worldly familiarity: ``A human has more common sense and knowledge regarding 
relations between entities and cause-and-effect patterns in the world than any single average machine.'' 

\item  Narrow specialties: ``A human's expertise in a given domain is expected to be more focused and limited than a machine's.'' For example, a human agent handling technical issues with the website of a health provider is not expected to also discuss actual medical issues. In general, machines do not have this limitation. 

\item  Learning: Turing claimed that humans retain both long and short term memory and learn from them, and machines often do not.  However, these days the opposite might be the case.  Machines can be equipped with vast memories, can access voluminous repositories of data, to which they can then apply powerful machine learning algorithms,where humans capabilities would be more limited. 
Still, one may say that ``humans can often be taught new tasks more simply than machines, sometimes by simple conversation''. For example, if a person needs an agent (human or machine) to carry out a task that 
the agent does not know how to do (in the case of a human agent)
or is not programmed for (in the case of a machine agent),
the person is more likely to succeed in teaching a willing human agent to carry out the task than in teaching a typical computerized agent.

\item  Collaboration: ``Machines are becoming better positioned to take advantage of multi-agent collaboration.'' For example, car-to-car communication for better coordination on a highway is probably easier to implement technologically than  establishing driver-to-driver communication. Thus the H-or-M question regarding a group of cars, wondering if they are all autonomous or are driven by multiple humans, may be partly answered by observing their coordination practices. The use of the idiom 'like a well-oiled machine' to describe the operation of an efficient human organization, hints at our intuition in this regard.

\item  Mistakes: ``Humans make more mistakes than machines.''

\item Diversity: ``Human behavior involves more randomness and arbitrary actions and is less predictable than that of machines.'' Thus, different humans working on the same task exhibit more diversity than different machines of the same model working on the same task. Similarly, the performance of a human repeating a given task is more diverse than that of a  machine repeating the task.  
\end{itemize}

Better understanding of these differences may pave the way to bridging them, by endowing machines with certain desired human capabilities, and to a lesser extent vice versa. This can be valuable not only for concealing the H-or-M answer in everyday tasks when appropriate, but for improving the performance of both humans and machines.  

\section {Some Aspects of H-or-M Interrogation}\label{sec:explicitImplicitInterrogation}

 As stated in the Introduction, in this paper we do  not seek 
 a strategy or protocol for eliciting the answer to the H-or-M question in everyday situations.
 Still, it is worthwhile to  discuss some relevant issues briefly, and to hint at a number of tentative ingredients of a potential strategy. 

In the classical Turing Test, only the interrogator is proactive, and they are 
in control of the interaction. The agent is expected to merely react to the inquiries and statements coming its way. Thus, this is an interrogation in the true sense of the word. 

Some variants of the Turing test are non-verbal in nature~\cite{ciardo2022TuringSimonTest,avraham2012roboticHandshake}. The interrogator challenges the agent to act in certain ways, and then analyzes the resulting behavior, including seeking patterns therein.  
However, in this case too the entire exchange is orchestrated as an interrogation. 

Related interrogations techniques can be found in Captcha challenges~\cite{vonAhn2003captcha} that are built around a cognitive task, and in human driven interrogations in a variety of contexts ranging from psychiatric therapy~\cite{heiser1979paranoyaSimulation} to reasoning about drone behavior~\cite{traboulsi2021HorMforDrone}.

Here, we turn our attention to situations where (i) the agent does not offer a direct answer to the H-or-M question, and (ii) the interrogation is passive, or implicit. For example, when a person is interacting with a service center the conversation is expected to be focused on the service issue at hand, rather than on unmasking the agent's H-or-M identity. If the person is interested in this information, they should derive it  from the agent's communications on the service issue only. 
Similarly, a human driver observing the non verbal
behavior of a nearby vehicle might  be interested in determining whether it is autonomous or driven by another human. This determination would normally be carried out through a combination of passive and uninvolved observation and ordinary road behavior, such passing the vehicle in question or getting close to it.

This leads to another aspect of interrogation: how does the person interact with the agent? Clearly, even just seeing the agent in action may provide 
some relevant clues. 
Hearing is another important channel. The classical Turing test is constrained to typewritten  textual interaction. However, while this limitation seems appropriate for achieving fairness --- since it masks gender differences between human speakers and overcomes technological constraints in speech synthesis --- it robs the interrogation of emotional aspects manifested in speech prosody. 
This could  be appropriate for testing  intelligence with less of a focus on emotions, but it may be inappropriate if we are interested in the H-or-M question in  interactions that normally involve speech. The same may also apply to interactions where agent actions could involve touch, smell, and possibly even taste.

What about other kinds of physical interactions? Can the interrogator ask to see the results of a blood test of the agent? 
We leave this aspect and other ``limitations of imitation'' for a future discussion. 

With the fast improvement in machine capabilities and usage, we expect the importance of the H-or-M issue to increase over time. Thus, it is likely that techniques will evolve for conducting proactive interrogations in a concealed manner, within  matter-of-fact verbal or non-verbal interactions. In some cases, these techniques may be crafted from the knowledge we have about distinctions between human and machine behavior, and in others they  may evolve naturally or subconsciously, and may even lead to better understanding of differences between humans and machines.  Such techniques may be supported by sharing historical information about interactions and interrogation results. 

The ability to conduct productive implicit H-or-M interrogations may even become an algorithmic/computational thinking skill, or perhaps a social skill, that humans will be expected to acquire. Furthermore, if the techniques can be formalized and generalized, we may see automated tools that help humans, as well as other automated agents, in conducting such delicate interrogations. And, if  such interrogations  can be formalized or automated, will humans and machines eventually learn to detect them? Such detection could trigger direct responses, in order to save time and effort, or perhaps drive redoubled efforts to conceal the answer. Would a human agent be offended if they notice that the person they are interacting with is not sure that they are indeed human?  Will people use such interrogation to tease agents, or perhaps to hint that the agent's behavior is too rigid? 

Finally, it is possible that while the H-or-M matter will be highly relevant, 
no specific effective interrogation protocols will evolve in the foreseeable future. 
In fact, social norms or judicial regulations may result in a practice of routinely disclosing an agent's H-or-M identity.
Moreover, in some contexts, people may just learn to live with not knowing and not asking---as is the case with determining the gender of one’s counterpart in a text-only interaction, although this is known to be a primary component of first impressions~\cite{signorella1992rememberingGender,ruble1986stalkingGender}.

Technological deficiencies in mimicking humans may render the entire issue moot,
and, conversely, technological superiority over human performance in key aspects of the interaction
may cause developers to forego the effort to mimic humans in secondary aspects.
Human agents in roles that are also fulfilled by machines may limit their own behaviors to purely professional and bureaucratic,
thus mimicking machines and reducing the advantages (or the significance of the differences) of interacting with a human.
Conversely, humans in such roles may emphasize behaviors that disclose their being human.
Finally, it is possible that while interrogation protocols will be developed, both humans and machines will learn to detect them and avoid playing along, rendering the protocols useless.

\section{Discussion}\label{sec:discussion}
While the issues and questions we have raised regarding the Human-or-Machine matter may pique one's curiosity, 
we may still ask, why are they interesting now? 
Why do we want to know now what people will do with answers to the Human-or-Machine question in common interactions? Can't we
just wait and see? Why do we care about what will people do when they find out that the agent they thought was a human, was really a machine, or vice versa? 
Why do we need to articulate differences between humans and machines with regard to observable behaviors in certain contexts? 

We believe that better understanding of these issues can advance science and technology in many ways.  Here are some examples. 

First, in current methods for developing computerized agents, the design of human interfaces involves a delicate balance between the value of friendly, intuitive human-oriented behavior (say, through the use of natural language), and the value of succinctness and predictability of typical machine behavior (say, using menu-based selections). Understanding  how human behaviors and expectations differ between interacting with humans and  machines can help in the design of human-computer interfaces and business processes. This may improve agent development productivity and the quality of the final products, and could also contribute to the satisfaction and even the well-being of human users. 

For example, if it turns out that people use a different subset of the language when interacting with machines that understand natural language, the training of such agents may become more focused and more efficient than when training on general natural language, and lead to fewer misunderstandings in real-world interactions. 

Second, a major factor in rich interactions is trust. Understanding the  differences between how trust-building emerges in human-human interaction as compared to human-machine interaction may allow us to better understand this elusive concept and to create protocols for enhancing and accelerating trust-building in a variety of contexts.   

Third, we are all familiar with cartoons depicting people grumbling or getting angry with their computers. 
For our own well-being, knowing that we are interacting with a machine rather than with a human may require us to channel our own 
natural emotions differently.
System developers are already well aware of the fact that certain system behaviors may evoke anger,  frustration, and other emotions. Translating such knowledge into system design decisions will  become even more complicated in the case of agents that mimic humans.
While there is a body of research about various aspects of human emotions in the context of
interactions with chatbots, the challenge here may be broader, both in terms of the different kinds of agents, and in connection to the very fact that a growing portion  of one's interactions may eventually be carried out with
machines. 
Research and therapy methods related to this area are emerging~~\cite{chaturvedi2023socialCompanionAI,heiser1979paranoyaSimulation}. 

Fourth, the design of machine-machine interfaces may change in various ways. 
On a basic level, two machine agents may use natural language conversations to discover each other's programming interfaces and available data (mimicking an engineer studying the documentation of such interfaces), and then switch to more efficient or more extensive interactions.
Such flexible interfaces will manifest enhancements over technologies, such as those of web sites that are designed for human use but also offer  special information and interfaces for web crawlers associated with search engines, or of registry-based service discovery in service-oriented architectures (SOA). We might see human-like agents that are interested in the H-or-M identity of whoever they interact with, and based on the answer will then act differently, or refrain from certain interactions altogether, insisting on interacting only with humans, or only with certain machines. 
More advanced interfaces may exhibit new self-organizing multi-agent collaboration, where  agents inquire about each other's broader goals and learning capabilities and then collaborate accordingly.
   
Carrying out research on human interaction with 
agents who mimic human behavior with high fidelity 
in common, real world situations
may not be easy at all. Will researchers be able to create the everyday nature of such interactions in a controlled environment? 
Will lab experiments with a limited number of kinds of machine agents be representative? 
And conversely, when collecting data from real-world interactions, will enough ground truth information be available with regard to whether the agents are humans or machines?  

In summary, recent technological advances give intelligent machines critical roles in our everyday lives. We do not know if such machines will come to be treated as conventional objects, like personal computers or ATMs, or as different kinds of living species, or perhaps, in the long run and in particular cases, even become indistinguishable from human professionals. 

However that may turn out, we are convinced that determining whether one is interacting with a machine or with another human is likely to become a central question. The insights to be gained from studying the question and its ramifications may have surprised even Turing. 

\section*{Acknowledgements}
The authors thank Joseph Sifakis for valuable discussions and suggestions.
This research was funded in part by an NSFC-ISF grant issued jointly by the  National Natural Science Foundation of China (NSFC) and the Israel Science Foundation (ISF grant 3698/21). Additional support was provided by a research grant from the Estate of Harry Levine, the Estate of Avraham Rothstein, Brenda Gruss, and Daniel Hirsch, the One8 Foundation, Rina Mayer, Maurice Levy, and the Estate of Bernice Bernath.

\section*{Declarations}
The authors contributed equally to the paper. 

\noindent The authors have no competing interests.

\bibliographystyle{apalike}

\end{document}